\documentclass[11pt]{article}

\usepackage[preprint]{acl}

\usepackage{times}
\usepackage{latexsym}

\usepackage[T1]{fontenc}

\usepackage[utf8]{inputenc}

\usepackage{microtype}

\usepackage{inconsolata}

\usepackage{graphicx}

\usepackage{ragged2e}
\usepackage[table,svgnames,HTML]{xcolor}
\usepackage[normalem]{ulem}
\usepackage[most]{tcolorbox}

\usepackage{booktabs}
\usepackage{multirow}
\usepackage{arydshln}

\usepackage{amsmath}
\usepackage{amssymb}
\usepackage{sidecap}
\usepackage{pifont}
\usepackage{enumitem}

\usepackage{url}

\definecolor{green1}{HTML}{E1EEBC}
\definecolor{green2}{HTML}{90C67C}
\definecolor{green3}{HTML}{67AE6E}
\definecolor{green4}{HTML}{A4B465}
\definecolor{green5}{HTML}{DCEAB6}

\definecolor{gray2}{HTML}{AEAEAE}
\definecolor{gray1}{HTML}{F0F0F0}
\definecolor{titlebg}{HTML}{111111}
\definecolor{clippedGray}{HTML}{E0E0E0} 
\definecolor{textGray}{HTML}{999999}    

\definecolor{weightHigh}{HTML}{D9534F} 
\definecolor{weightMed}{HTML}{E67E22}  

\tcbset{
  tokenpill/.style={
    on line,
    boxrule=0pt,
    arc=2.2pt,
    left=2.2pt,right=2.2pt,top=1.2pt,bottom=1.2pt,
    boxsep=0.2pt,
    coltext=black,
  },
  clippedpill/.style={
    tokenpill,
    colback=gray1,
    coltext=black,
  },
}

\newtcbox{\TokPill}[1][]{tokenpill,#1}
\newtcbox{\ClipPill}[1][]{clippedpill,#1}

\newtcolorbox{box2}[2][]{%
  float*=t,
  width=\textwidth,
  enhanced,
  colback=white,
  colframe=black,
  boxrule=0.6pt,
  arc=3pt,
  left=8pt,right=8pt,top=6pt,bottom=6pt,
  fonttitle=\bfseries\color{white},
  coltitle=white,
  attach boxed title to top left={xshift=0.0mm,yshift=-2.0mm},
  boxed title style={
    colback=titlebg,
    colframe=titlebg,
    arc=3pt,
    boxrule=0pt,
    left=8pt,right=8pt,top=4pt,bottom=4pt,
  },
  code={\refstepcounter{table}},
  title={Table~\thetable: #2},
  #1
}

\newcommand{\hltoken}[2]{%
    \begingroup
    \setlength{\fboxsep}{1.5pt}
    \colorbox{#1}{\strut #2}%
    \endgroup
}

\newcommand{\hltokens}[2]{%
    \begingroup
    \setlength{\fboxsep}{1.15pt}
    \colorbox{#1}{\strut #2}%
    \endgroup
}

\newcommand{\notedtoken}[3]{%
    \begingroup
    \setlength{\fboxsep}{1.5pt}%
    \colorbox{#1}{\strut #2$_{\text{#3}}$}%
    \endgroup
}

\newtcolorbox{HeadlessBox}[1][]{%
    enhanced,
    colframe=black!80,   
    colback=white,       
    boxrule=0.8pt,       
    arc=1.5mm, auto outer arc, 
    boxsep=4pt,
    left=8pt, right=8pt, top=6pt, bottom=8pt,
    fontupper=\RaggedRight\linespread{1.4}\selectfont, 
    #1
}

\newtcolorbox{box3}[2][]{%
    width=\textwidth,      
    enhanced, fonttitle=\bfseries\large,
    breakable,
    colframe=black!80,     
    colback=white,         
    colbacktitle=black!90, 
    boxrule=0.8pt,         
    arc=1.5mm, auto outer arc,
    boxsep=4pt,            
    left=8pt, right=8pt, top=6pt, bottom=8pt, 
    title={#2}, 
    code={\refstepcounter{table}}, 
    fontupper=\RaggedRight\linespread{1.4}\selectfont, 
    #1 
}

\newtcolorbox{box1}[2][]{%
  float*=t,              
  width=\textwidth,      
  enhanced,
  colback=white,
  colframe=black,
  fonttitle=\bfseries,
  code={\refstepcounter{table}}, 
  title={Table~\thetable: #2},
  #1
}

\title{QaRL: Rollout-Aligned Quantization-Aware RL for Fast and Stable Training under Training--Inference Mismatch}

\author{
\textbf{Hao Gu$^{1}$}
\quad
\textbf{Hao Wang$^{2}$}
\quad
\textbf{Jiacheng Liu$^{1}$}
\quad
\textbf{Lujun Li$^{1}$}
\quad
\textbf{Qiyuan Zhu$^{1}$}
\\
\textbf{Bei Liu$^{1}$}
\quad
\textbf{Binxing Xu$^{3}$}
\quad
\textbf{LEI WANG$^{1}$}
\quad
\textbf{Xintong Yang$^{1}$}
\quad
\textbf{Sida Lin$^{1}$}
\quad
\textbf{Sirui Han$^{1}$\thanks{Corresponding authors.}}
\quad
\textbf{Yike Guo$^{1}$\footnotemark[1]}
\\[0.5em]
$^{1}$The Hong Kong University of Science and Technology
\\
$^{2}$City University of Hong Kong
\qquad
$^{3}$Zhejiang University
\\
\texttt{marcusguhao@gmail.com}
\qquad
\texttt{siruihan@ust.hk}
\qquad
\texttt{yikeguo@ust.hk}
}

\begin{document}
\maketitle
\begin{abstract}
Large language model (LLM) reinforcement learning (RL) pipelines are often bottlenecked by rollout generation, making end-to-end training slow. Recent work mitigates this by running rollouts with quantization to accelerate decoding, which is the most expensive stage of the RL loop. However, these setups destabilize optimization by amplifying the training–-inference gap: rollouts are operated at low precision, while learning updates are computed at full precision. To address this challenge, we propose QaRL (Rollout Alignment Quantization-Aware RL), which aligns training-side forward with the quantized rollout to minimize mismatch. We further identify a failure mode in quantized rollouts: long-form responses tend to produce repetitive, garbled tokens (error tokens). To mitigate these problems, we introduce TBPO (Trust-Band Policy Optimization), a sequence-level objective with dual clipping for negative samples, aimed to keep updates within the trust region. On Qwen3-30B-A3B MoE for math problems, QaRL outperforms quantized-rollout training by +5.5 while improving stability and preserving low-bit throughput benefits.
\end{abstract}

\section{Introduction}
Recent reasoning LLMs, such as OpenAI o1~\citep{jaech2024openai}, have demonstrated the effectiveness of Chain-of-Thought (CoT)~\citep{wei2022chain} for solving complex problems. More recently, DeepSeek-R1~\citep{guo2025deepseek} has shown that reinforcement learning with simple rule-based reward functions  (RLVR), can induce emergent reasoning behaviors and yield gains in challenging domains such as math problem solving~\citep{deepscaler2025}. A key driver behind these improvements is test-time scaling~\citep{muennighoff2025s1}: reasoning models often generate longer CoT, trade additional computation for accuracy. However, longer generations increase training-time cost, since RL continously samples long responses during optimization.

\begin{figure}
    \centering
    \includegraphics[width=1\linewidth]{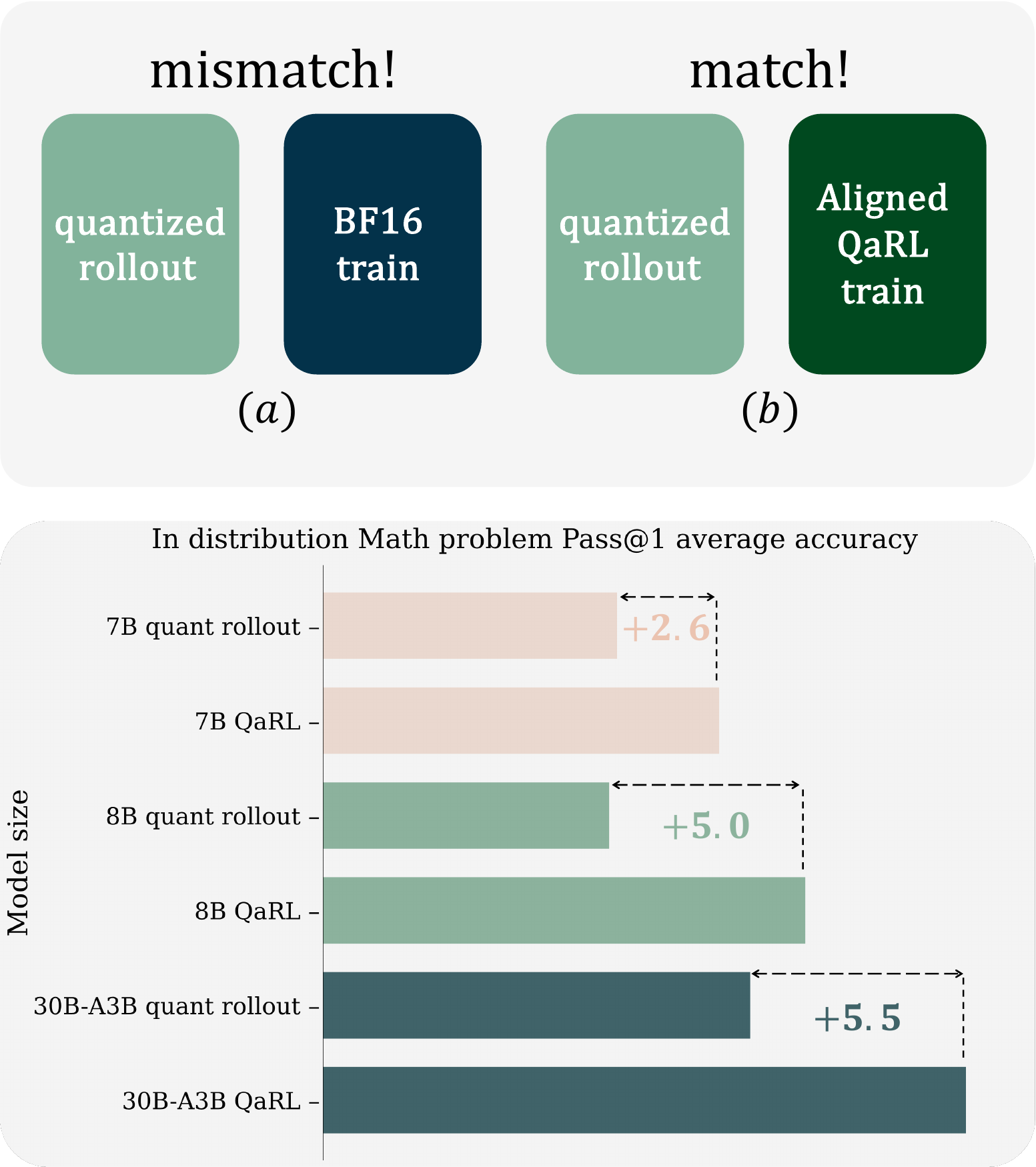}
    \caption{(a) Quantized rollout alone introduces mismatch. (b) Aligned QaRL alleviates it, improving performance across model sizes, including MoE.}
    \label{fig:intro}
    \vspace{-6mm}
\end{figure}
Unlike SFT, which requires only a single forward pass, a standard RL step~\citep{sheng2024hybridflow,openrlhf} for LLMs involves three phases: (i) rollout to generate responses, (ii) a forward pass to compute token probabilities, and (iii) a backward pass to update model via policy gradients. Autoregressive decoding generates response token by token, making rollouts the dominant cost, comprising roughly 70\% of RL training time. Consequently, quantizing rollout model is a natural way to accelerate RL optimization~\citep{liu2025flashrl}. However, it also creates a training-inference mismatch challenge: responses are sampled from low bit rollout, while updates are performed by full precision learner.

To addresses both efficiency and stability, in this work, we study \textbf{quantization-aware RL}. By executing rollouts under low-bit quantization, we substantially reduce the dominant generation cost and accelerate end-to-end RL training. To mitigate the resulting training-inference mismatch, we further perform rollout-aligned quantization-aware training on the learner side, aligning learner’s policy with the quantized behavior used for sampling.

Moreover, we find a critical failure mode in quantized rollouts: noise accumulates over long generations, producing off-trajectory repetitive and garbled tokens. These \textbf{error tokens} are typically assigned very low probability under the policy, driving the policy ratio (and mismatch reweighting) to extreme. Such outliers are not reliably controlled by standard PPO-style clipping, breaking the intended trust region behavior and destabilizing training. To address this, we introduce a trust-band control: we apply dual clipping tailored to negative samples and perform sequence-level objectives on both the policy ratio and the mismatch weight, drop entire responses that exceed the bands.

Extensive experiments on math, logic and code benchmarks demonstrate that QaRL matches BF16 training performance and outperform quantized rollout training. Even on unstable MoE models such as Qwen3-30B-A3B-Base, QaRL achieves an average math score of 51.2, close to BF16's 52.1 and surpassing quantized rollout's 45.7, while delivering a 1.3$\times$ training speedup over BF16. Our contributions are as follows:
\begin{enumerate}[leftmargin=*, itemsep=1pt, topsep=2pt, parsep=0pt, partopsep=0pt]
    \item We present a practical quantized aware RL pipeline for decoupled, hybrid RL systems, and introduce \textbf{QaRL} to minimize the mismatch between quantized rollouts and learner-side training.
    \item We identify error tokens as the key cause of training instability under quantized rollouts, and propose \textbf{Trust Band Policy Optimization (TBPO)}, a sequence-level dual-clipping strategy that keeps updates within a trust region and stabilizes convergence.
\end{enumerate}



\section{Preliminaries}
\subsection{Group Relative Policy Optimization.}
GRPO~\citep{shao2024deepseekmath} is a PPO style~\citep{schulman2017proximal} policy gradient objective that removes the value function (critic) and uses group relative rewards as the baseline. Given a query $q$, we sample a group of $G$ responses $\{o_1,\ldots,o_G\}$ from the old policy $o_i\sim \pi_{\theta_{\text{old}}}(\cdot\mid q)$. Let $r_i$ be the scalar (verifiable) reward for $o_i$ and $|o_i|$ its length. GRPO defines a group-normalized advantage $A_i$ (broadcast to tokens as $A_{i,t}=A_i$) and the token-level importance ratio $R_{i,t}(\theta)$:
\begin{equation*}
\begin{aligned}
A_{i}&=\frac{r_{i} - \mathrm{mean}\{r_1,\ldots r_G\}}{\mathrm{std}\{r_1,\ldots r_G\}}, \\
&R_{i,t}(\theta)=\frac{\pi_\theta(o_{i,t}\mid q_i)}{\pi_{\theta_{\text{old}}}(o_{i,t}\mid q_i)}
\end{aligned}
\end{equation*}
To prevent overly large updates and ensure the update remains within the trust region~\citep{schulman2015trust}, GRPO applies PPO-style clipping and defines the token-level clipped policy loss:
\begin{equation*}
\mathcal{L}_{i,t}(\theta)
= -\min\!\left(R_{i,t}(\theta)\,A_i,\;\tilde R_{i,t}(\theta)\,A_i\right),
\end{equation*}
where $\tilde R_{i,t}=\mathrm{clip}(R_{i,t},1-\epsilon,1+\epsilon)$ and $\epsilon$ controls the trust region. The GRPO optimization objective is defined as follows:
\begin{equation*}
\mathcal{L}_{\mathrm{GRPO}}(\theta)
=\mathbb{E}_{q,\,o\sim \pi_{\theta_{\text{old}}}}
\left[
\frac{1}{G}\sum_{i=1}^{G}\frac{1}{|o_i|}\sum_{t=1}^{|o_i|}\mathcal{L}_{i,t}(\theta)
\right].
\end{equation*}

\subsection{Quantization}
Quantization accelerates LLM inference by enabling low-bit GEMM operations, where weights and activations are compressed into low bit representations that modern GPUs can process directly via Tensor Core kernels.

\noindent\textbf{Integer quantization.} A full-precision tensor $W\in\mathbb{R}$ is mapped to $W_q\in\mathbb{Z}$ via scale $s$ and zero-point $z$:
\begin{equation*}
\begin{aligned}
\textbf{Quant}&:\; W_q=\mathrm{clamp}\!\left(\left\lfloor \frac{W}{s}\right\rceil + z,\; q_{\min}, q_{\max}\right),\\
\textbf{Dequant}&:\; \hat W=s\,(W_q-z),
\end{aligned}
\end{equation*}
where $\lfloor\cdot\rceil$ denotes rounding. Quantized GEMM computes the integer matmul and rescales the result (accumulating in higher precision):
\begin{equation}
Y \approx s_x s_w \cdot \big((X_q-z_x)\,(W_q-z_w)\big).
\end{equation}

\noindent\textbf{Floating-point quantization.} For low-bit floating formats (e.g., FP8/FP4), values are cast to reduced precision with scaling:
\begin{equation*}
\begin{aligned}
\textbf{Quant}&:\; s=\frac{\max(|W|)}{\alpha},\; W_q=\big[\,W/s\,\big]_{\mathrm{FP}k},\\
\textbf{Dequant}&:\; \hat W=s\cdot W_q,
\end{aligned}
\end{equation*}
where $[\,\cdot\,]_{\mathrm{FP}k}$ denotes casting to $\mathrm{FP}k$ and $\alpha$ is the maximum finite representable value ($\alpha=6$ for e2m1FP4, $\alpha=448$ for e4m3FP8). Low-bit GEMM performs multiplication on quantized operands and rescales:
\begin{equation}
Y \approx (s_x s_w)\cdot\big(X_q W_q\big),
\end{equation}
with $X_q,W_q$ as the low-bit floating representations. We denote $\mathrm{x}$-bit weight and $\mathrm{y}$-bit activation quantization as W$\mathrm{x}$A$\mathrm{y}$.

\noindent \textbf{Quantization-Aware Training (QAT).} To recover accuracy lost in low-bit conversion, QAT integrates quantization noise into the training loop~\citep{liu2024llm}. Specifically, it injects \texttt{fake quant} into the forward pass to simulate rounding, while master weights remain in full precision. The Straight-Through Estimator (STE) approximates gradients for non-differentiable backward pass.

\noindent \textbf{Fully Quantized Training (FQT).} Distinct from QAT, which employs \texttt{fake quant} to merely simulate quantization noise while maintaining high-precision arithmetic, Fully Quantized Training (or Low-Bit Training) fundamentally executes operations using actual low-bit data types (e.g., FP8, INT8)~\citep{xi2024coat,zhao2025insights}. By leveraging low-precision hardware kernels for both forward and backward passes, this paradigm directly reduces the computational overhead and memory footprint of the training process itself.

\subsection{Training--Inference Mismatch in RL}
Modern LLM RL pipelines are typically hybrid systems: for throughput, rollouts are generated by a high-performance inference engine (e.g., vLLM, SGLang), while policy optimization are carried out by a training backend (e.g., FSDP/Megatron). Although both components load the same parameter $\theta_{\text{old}}$, the rollout and the training engine may produce different results for the same query ($\text{LLM}_{\text{rollout}}(q) \neq \text{LLM}_{\text{train}}(q)$). This occurs because they often implement different kernels (e.g., different attention kernels, batch variant operations). As a result, the rollout engine induces a \textbf{sampler policy} $\pi_{\text{sampler}}(\cdot\mid \theta_{\text{old}})$ that is not exactly the same distribution as the \textbf{learner policy} $\pi_{\text{learner}}(\cdot\mid \theta_{\text{old}})$. We refer to this discrepancy as \textbf{training--inference (learner--sampler) mismatch}.

Consider PPO's clipped objective:
\begin{equation*}
\label{eq:ppo-ideal}
\mathbb{E}_{a\sim \pi_{\theta_{\text{old}}}}
\Big[
\min\big(
r_\text{prox}\,\hat A, \mathrm{clip}(r_\text{prox},1-\epsilon,1+\epsilon)\,\hat A
\big)
\Big]
\end{equation*}
where proximal importance sampling $r_\text{prox}=\frac{\pi(a\mid\theta)}{\pi(a\mid\theta_{\text{old}})}$ and $\hat A$ is an advantage estimate. In a hybrid system, however, tokens are sampled from $\pi_{\text{sampler}}(\cdot\mid\theta_{\text{old}})$, while $r_\text{prox}$ is computed using learner's distributions:
\begin{equation*}
\begin{aligned}
\label{eq:ppo-mismatch}
\mathbb{E}_{a\sim \pi_{\textbf{sampler}}(\theta_{\text{old}})}
\Big[
\min\big(
\frac{\pi_{\textbf{learner}}(a\mid\theta)}{\pi_{\textbf{learner}}(a\mid\theta_{\text{old}})}\,\hat A,
\\
\mathrm{clip}(\frac{\pi_{\textbf{learner}}(a\mid\theta)}{\pi_{\textbf{learner}}(a\mid\theta_{\text{old}})},1-\epsilon,1+\epsilon)\,\hat A
\big)
\Big]
\end{aligned}
\end{equation*}
Consequently, the trust region is enforced on the sampler distribution rather than the learner distribution, causing the update to deviate from its intended target. Drawing inspiration from Decoupled PPO~\citep{hilton2022batch}, where a mismatch between the behavior and proximal policies is corrected via importance sampling, we can reweight the updates by multiplying with the ratio:
\begin{equation} 
\label{eq:tis-weight} 
w_\text{mismatch} = \frac{\pi_{\textbf{learner}}(a\mid\theta_{\text{old}})} {\pi_{\textbf{sampler}}(a\mid\theta_{\text{old}})} \end{equation}
Applying $w_{\text{mismatch}}$ yields
\begin{equation}
\begin{aligned}
\label{eq:ppo-tis}
\mathbb{E}_{a\sim \pi_{\text{sampler}}(\theta_{\text{old}})}&
\Big[
\frac{\pi_{\textbf{learner}}(a\mid\theta_{\text{old}})} {\pi_{\textbf{sampler}}(a\mid\theta_{\text{old}})}\cdot
\min\big(
r_{\text{prox}}\,\hat A,\\
&\mathrm{clip}(r_{\text{prox}},1-\epsilon,1+\epsilon)\,\hat A
\big)
\Big].
\end{aligned}
\end{equation}
Intuitively, this correction is conceptually orthogonal to PPO proximal ratio $r_\text{prox}=\frac{\pi_{\text{learner}}(a\mid\theta)}{\pi_{\text{learner}}(a\mid\theta_{\text{old}})}$, while $w_\text{mismatch}$ rectifies the distribution shift caused by the system-level training--inference mismatch. Together, they ensure policy optimization remains within the learner's trust region. Recent works have further explored this direction, such as \textbf{TIS}~\citep{yao2025offpolicy} (which truncates upper bound of $w_\text{mismatch}$) and \textbf{MIS}~\citep{liu-li-2025-rl-collapse} (reject samples with overly large $w_\text{mismatch}$ and apply sequence level $w_\text{mismatch}$).

\section{Methods}
\begin{figure}
    \centering
    \includegraphics[width=1\linewidth]{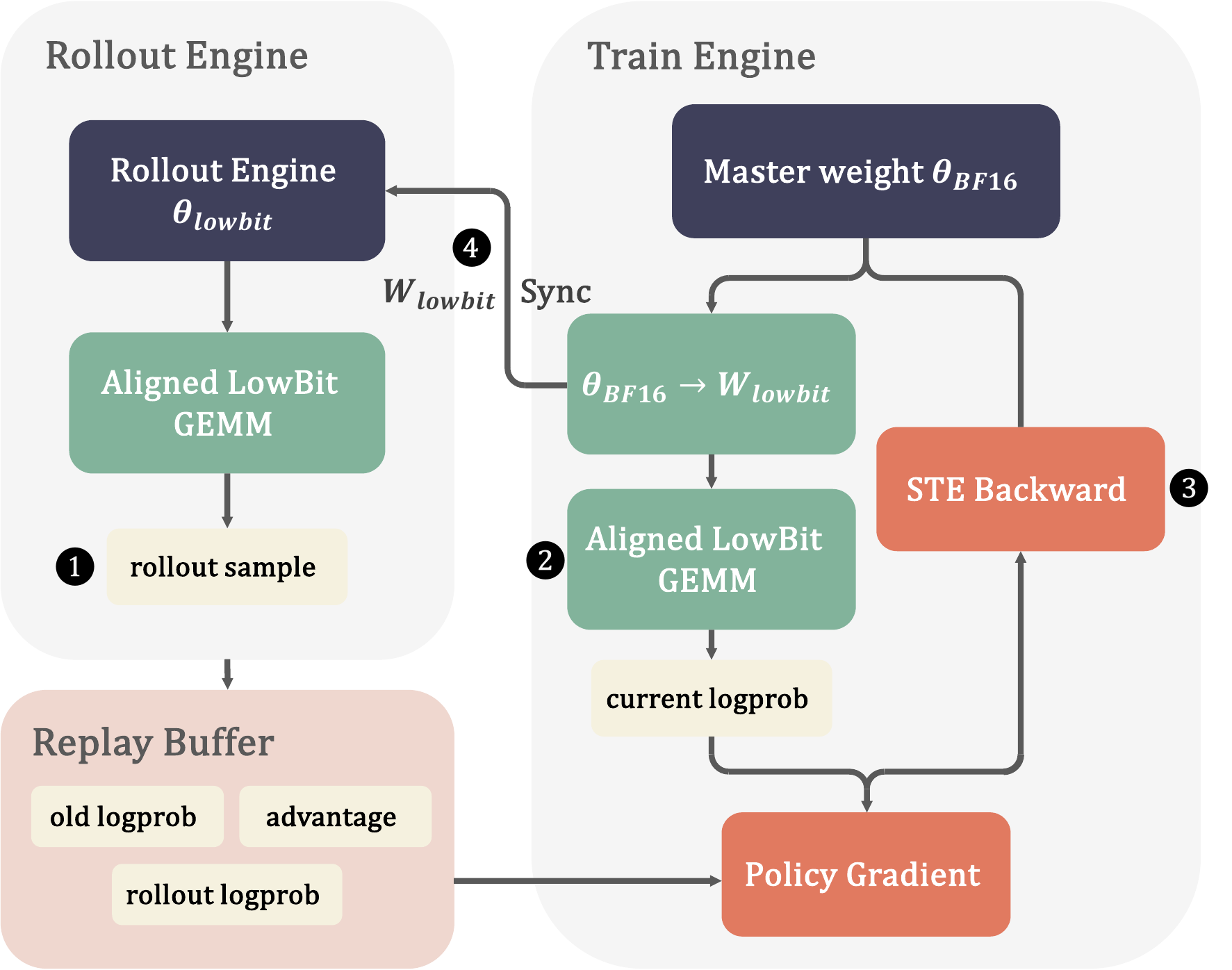}
    \caption{\textbf{Overview of the QaRL pipeline in a hybrid RL system.} \ding{182} The quantized rollout engine $\theta_\text{lowbit}$ generates samples. \ding{183} The training engine maintains $\theta_\text{BF16}$ master weights and performs rollout-aligned low-bit GEMM to compute current logprob. \ding{184} Policy gradients are computed using replay buffer data to update the model via STE. \ding{185} The updated low-bit weights $W_\text{lowbit}$ are synchronized to the rollout engine.}
    \label{fig:methods}
    \vspace{-3mm}
\end{figure}
\subsection{Rollout-Aligned Quantization Aware Reinforcement Learning}

To alleviate the decoding bottleneck during rollouts, a practical approach is to quantize the rollout model. By lowering the precision of weights and activations (e.g., W8A8 or W4A16), we can significantly accelerate rollout generation. However, quantizing only the rollout engine (we term this \textbf{Quantized Rollout Training}) introduces a more severe form of training--inference mismatch.

In this regime, the sampler policy $\pi_{\text{sampler}}$ becomes a quantized approximation $\pi_{\text{quant-sampler}}$, while the learner policy $\pi_{\text{learner}}$ typically remains full-precision $\pi_{\text{BF16-learner}}$. The resulting mismatch makes the importance weight
$$
w_{\text{mismatch}}=\frac{\pi_{\text{learner}}}{\pi_{\text{sampler}}}=\frac{\pi_{\text{BF16-learner}}}{\pi_{\text{quant-sampler}}}
$$
drift far away from $1.00$ as Fig.~\ref{fig:prob_ratio}(d). When both sampler and learner operate in full precision, the mismatch is typically limited to implementation details (e.g. different kernels), so the distribution shift is mild and can often be compensated by reweighting with $w_{\text{mismatch}}$. In contrast, quantized rollout training introduces larger distribution gap and $w_{\text{mismatch}}$ become more extreme. This effect is especially pronounced for long responses, quantization-induced errors accumulate across decoding steps, making the divergence grow with response length and increasingly difficult to correct. To fundamentally resolve this mismatch, we align the learner with the quantized sampler by adopting \textbf{Quantization-Aware Reinforcement Learning (QaRL)}.

But building a quantized training pipeline in a decoupled hybrid RL stack is non-trivial: rollouts are generated by an inference engine, while policy optimization runs in a separate training backend. To address this challenge, we decompose our implementation into three components: \textbf{(1) low-bit rollout inference}, \textbf{(2) rollout aligned quantization aware training}, and \textbf{(3) weight synchronization from training to inference}. Concretely:

\textbf{1. Quantized inference in the rollout engine.} We deploy a low-bit rollout model in the inference engine to accelerate decoding. Empirically, for larger models (especially \textbf{MoE}), we observe that W4A16 can deliver higher throughput than W8A8 due to memory constraints. Therefore, we support both configurations to maximize rollout efficiency across different model regimes.

\textbf{2. Rollout aligned quantization aware training.} On the training side, we maintain master weights $\theta_{\text{BF16}}$ in high-precision. Standard QAT simulates quantization error by inserting fake-quant operators $\big[\text{dequant}[\text{quant}(\theta_{\text{BF16}})] \big]$, while computing in high precision. To minimize the gap between $\pi_{\text{quant-sampler}}$ and $\pi_{\text{quant-learner}}$, we perform on-the-fly quantization during the forward pass:
$$\hat{\mathbf{W}}_{\text{low-bit}} = \text{Quant}(\theta_{\text{BF16}})$$
Critically, rather than merely simulating quantization, we execute low-bit GEMM directly on these quantized tensors, thereby precisely mirroring the arithmetic behavior of the rollout engine. During the backward pass, gradients are computed via the Straight-Through Estimator (STE) and applied to the master weights $\theta_{\text{BF16}}$.
Our approach lies between fake quant QAT and end-to-end low-precision FQT: we execute the forward pass in low bit while keeping the backward pass in full precision. We avoid FQT for larger quantization error introduced by 4-bit gradients.

\textbf{3. Weight updates from training to inference.} After each optimization step, the learner’s parameters change, requiring the rollout engine to be refreshed accordingly. Since the training engine already materializes $\hat{\mathbf{W}}_{\text{low-bit}}$ for the aligned low-bit GEMM, we directly publish these low-bit tensors to the inference engine at each step. This avoids redundant re-quantization and ensures the sampler weights remain in the exact low-bit format learner uses during forward. The overall pipeline is illustrated in Fig.~\ref{fig:methods}.

Notably, unlike prior work~\citep{vllm2025bitwise,sglang2025deterministic} that enforces strictly bitwise consistent on-policy execution, our approach tolerates discrepancy between the sampler and learner policies $\pi_{\text{quant-sampler}}\neq\pi_{\text{quant-learner}}$ and compensates for it using $w_\text{mismatch}$. Achieving bitwise alignment requires both batch and tensor-parallel invariance kernel, which is not a "free lunch", incurring an average 2$\times$ slower~\citep{he2025nondeterminism,zhang2025deterministicinferencetensorparallel}. Consequently, rather than bitwise-identical kernels, we focus on using aligned low-bit forward to ensure that optimization remains robustly within the intended trust region. As for master weight precision, previous work~\citep{qi2025defeating} claims that using the finer granularity of $\theta_\text{FP16}$ can mitigate mismatch. However, we empirically find that using $\theta_\text{FP16}$ with dynamic loss scaling causes gradient NaN underflow, and this phenomenon does not occur with $\theta_\text{BF16}$.

\begin{figure}
    \centering
    \includegraphics[width=1\linewidth]{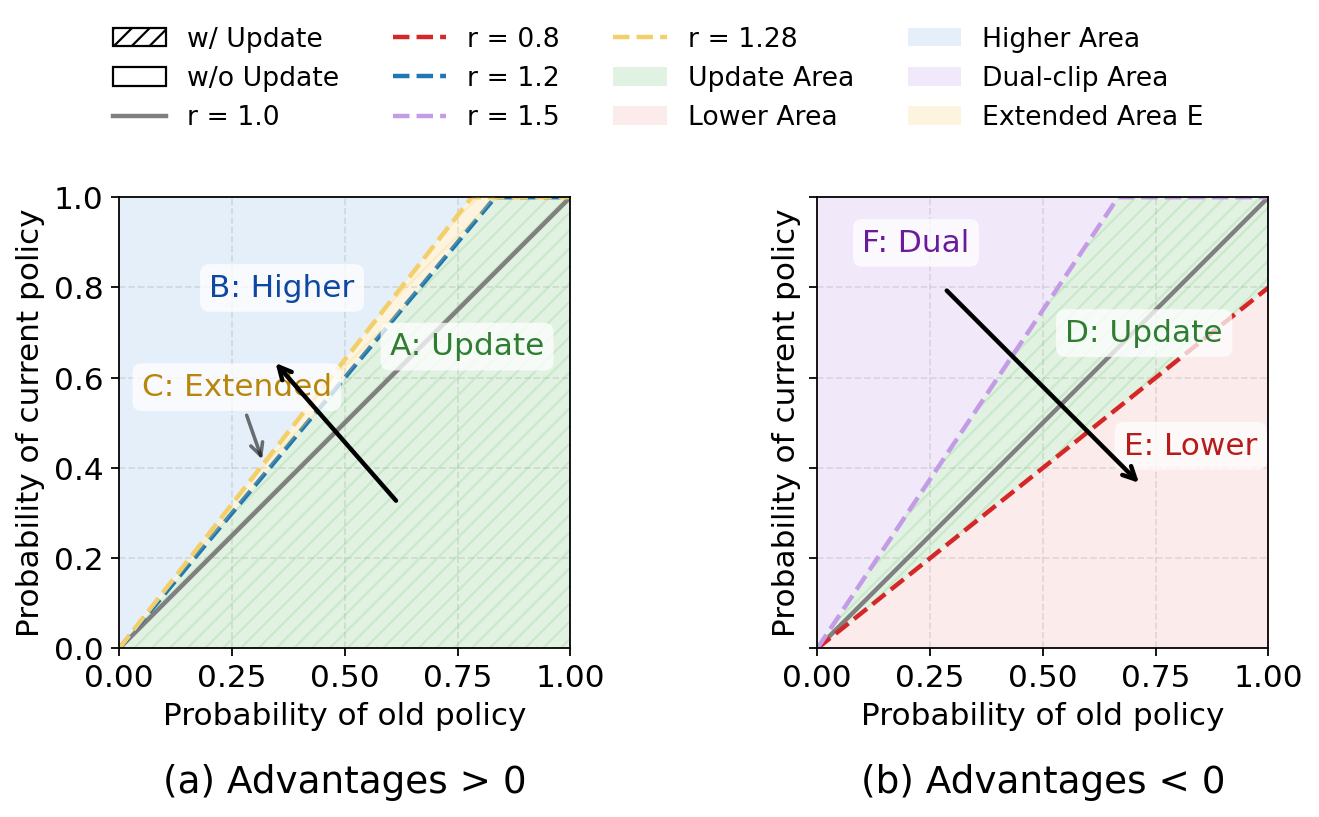}
    \caption{\textbf{Token level policy clipping regions.} Axes represent token probabilities under the old and current policies, with the slope defining the $r_\text{prox}=\text{prob}_\text{current}/\text{prob}_\text{old}$, with arrows indicating the direction of the policy update.}
    \label{fig:clip_region}
    \vspace{-3mm}
\end{figure}
\subsection{Trust Band Policy Optimization}

Although rollout-aligned quantized training substantially mitigates instability arising from training--inference mismatch, we observe a remaining failure mode: under-trained quantized policy tend to produce \textbf{repetitive and garbled tokens} in long responses, which we term \textbf{error tokens}, illustrated in appendix table~\ref{app:tokens}. This stems from the error-amplification dynamics of autoregressive decoding: an error token at step $t$ sends the model off-trajectory, causing subsequent tokens generated from a corrupted state and amplifying degradation.

\begin{figure}
    \centering
    \includegraphics[width=1\linewidth]{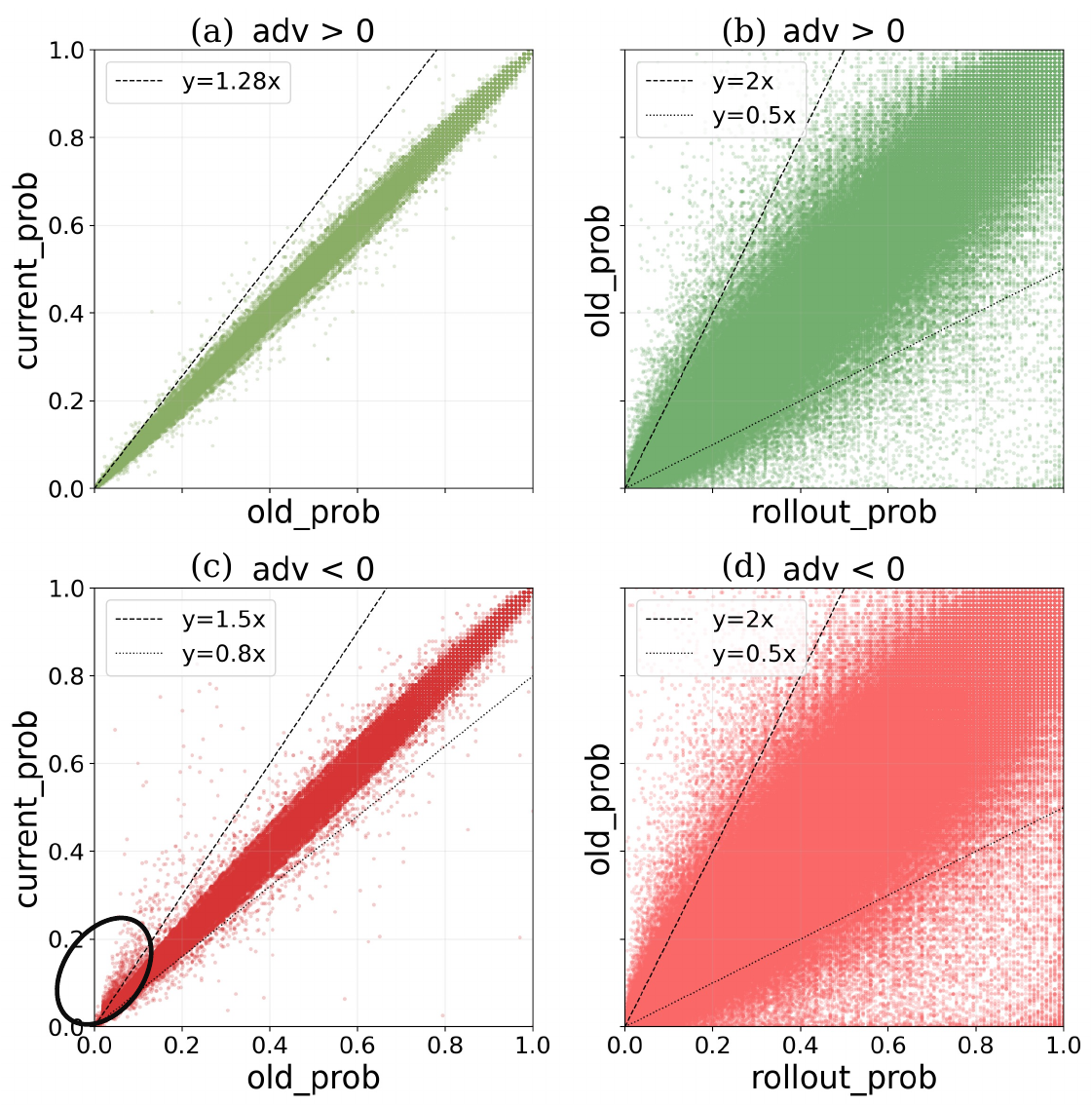}
    \caption{\textbf{Distribution of token level $r_\text{prox}$ and $w_\text{mismatch}$.} Lines indicate clipping boundaries.}
    \label{fig:prob_ratio}
    \vspace{-3mm}
\end{figure}

This phenomenon interacts closely with the trust-region control in PPO-style objectives. Standard PPO clipping operates directionally: for positive advantages, it constrains only the upper bound to $(0, 1+\epsilon)$, while for negative advantages, it constrains only the lower bound to $(1-\epsilon, +\infty)$. These prevent the updated policy drift too far from the old policy in the update direction. Recently, several works have revisited this design. For example, DAPO~\citep{yu2025dapo} proposes clipping higher for $A>0$ as \texttt{C:extended} region in Fig.~\ref{fig:clip_region}(a), arguing that high-entropy which correspond to low-probability tokens, can encourage exploration.

\begin{figure}
    \centering
    \includegraphics[width=1\linewidth]{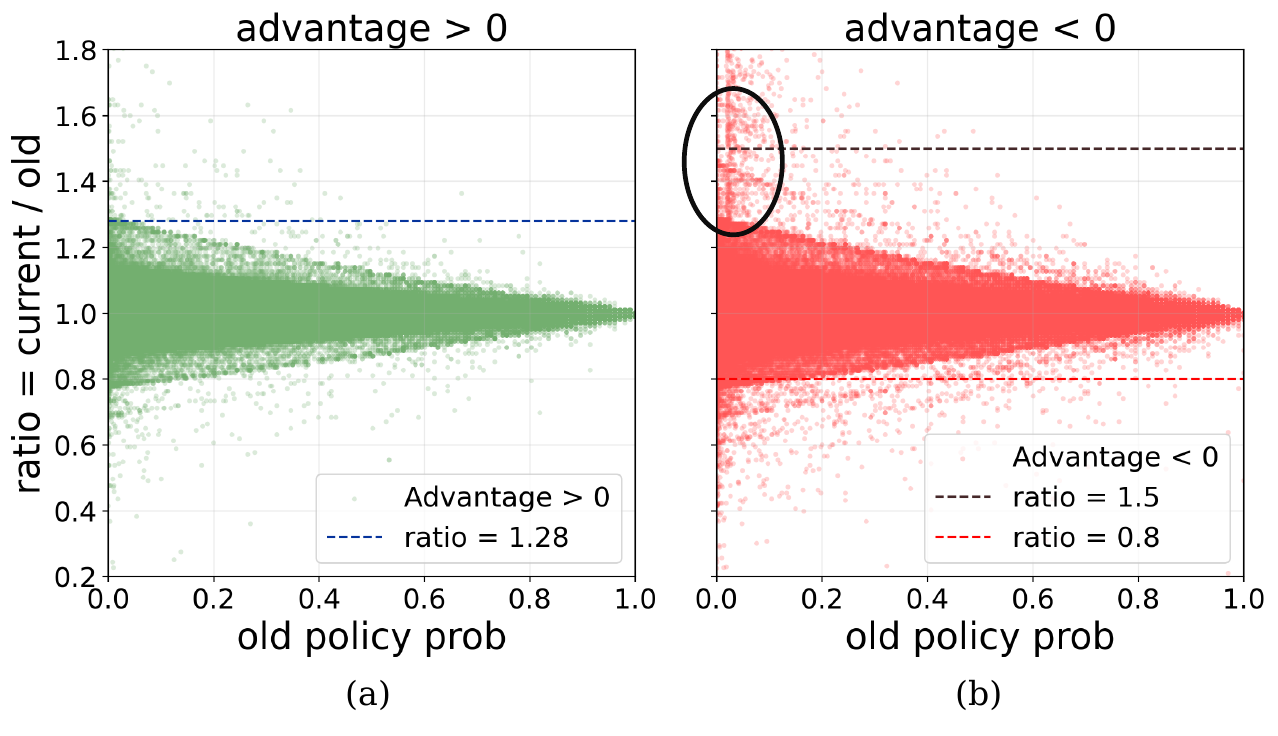}
    \caption{\textbf{Relationship between $r_\text{prox}$ and old policy probability.} Low-probability negative tokens are more prone to extreme $r_\text{prox}$.}
    \label{fig:is_region}
    \vspace{-4.5mm}
\end{figure}

\textbf{Dual Clipping.} In our setting, the more critical issue emerges with negative samples. Empirically, error tokens are rare in positive trajectories but predominate in negative ones. As shown in Fig.~\ref{fig:prob_ratio}(c), these negative error tokens are concentrated in the low-probability regions of both the old and current policies ($\pi_{\text{old},t}$ and $\pi_{\theta,t}$). In this regime, the PPO policy ratio becomes more sensitive to baseline probabilities:

{\small
\begin{align*}
r_{\text{prox},t}=\begin{cases} 
0.80/0.77\approx1.03 & p_{\text{old}}=0.77,\,p_{\theta}=0.80\\  
0.05/0.02=2.5 & p_{\text{old}}=0.02,\,p_{\theta}=0.05
\end{cases}
\end{align*}
}
Here, identical absolute perturbations produce dramatically different relative changes depending on $\pi_{\text{old},t}$. For typical high-confidence tokens (e.g., $p_{\text{old}}=0.77 \rightarrow p_{\theta}=0.80$), the ratio remains modest at $\approx1.07$. Conversely, for low-probability tail tokens (e.g., $p_{\text{old}}=0.02 \rightarrow p_{\theta}=0.05$), the ratio explodes to $2.5$. And low-prob tokens are inherently more prone to probability shifts. This effect is directly visible in Fig.~\ref{fig:is_region}(b): when $\pi_{\text{old}}$ is small, negative-advantage tokens exhibit a much heavier tail of $r_{\text{prox}}$, producing extreme ratios. Under $A_t<0$, standard PPO clipping is one-sided (it only enforces a lower bound), so these high $r_{\text{prox}}$ tail tokens can dominate the gradient magnitude, inflate variance, and break the intended trust region. To mitigate this failure mode, we leverage \textbf{a negative dual clipping scheme}~\citep{ye2020mastering}: for $A<0$, we not only enforce the lower bound but also impose an explicit upper bound as \texttt{F:Dual} region in Fig.~\ref{fig:clip_region}(b), preventing exploding ratios from low-probability error tokens from dominating the update.

\begin{figure}
    \centering
    \includegraphics[width=1\linewidth]{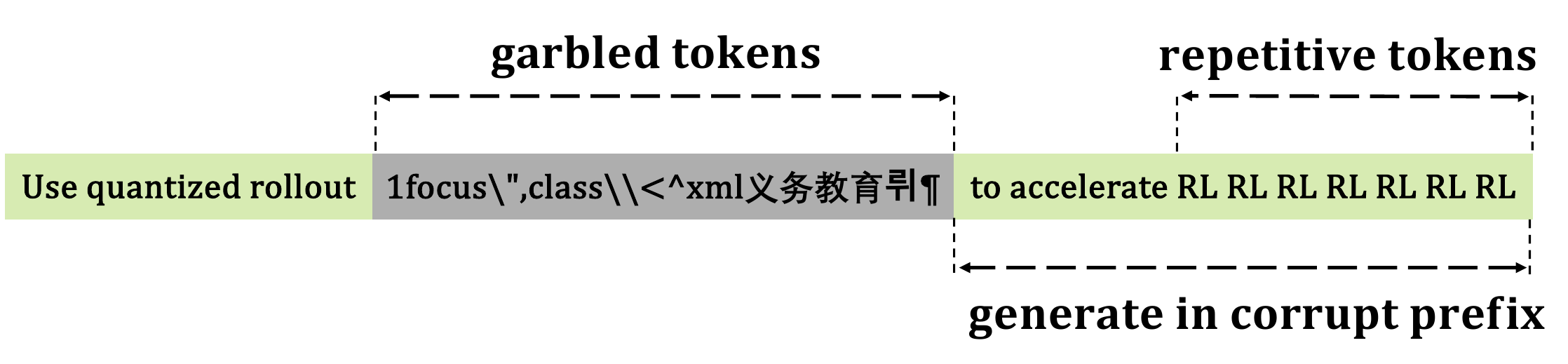}
    \caption{\textbf{A mid response error propagates to future tokens.} Although the initial \hltokens{gray2}{garbled tokens} are clipped, the \hltokens{green5}{repetitive tokens} induced by this error are not clipped by token-level objectives.}
    \label{fig:error_token}
    \vspace{-4mm}
\end{figure}

\textbf{Sequence Level Objectives.} Furthermore, since RLVR optimize outcome rewards, GRPO's token-level importance sampling serves merely as a first-order approximation. Token-level clipping fails to rectify \textbf{mid-response errors}: once a single token deviates, the subsequent tokens no longer lie on the intended trajectory (Fig~\ref{fig:error_token}).

To remedy this, we return to a stricter trust-region view and treat each entire response as single action. Concretely, we apply sequence-level ratios and clipping~\citep{zheng2025group}, mask out the entire responses from policy update when their sequence-level ratio exceeds the bounds. 
Another advantage of sequence-level ratio is its ability to discriminate error tokens and low-prob exploration tokens, which token-level bounds cannot resolve. Error tokens typically occur in clusters, causing high variance in $r_{\text{seq-prox}}$ that readily exceeds trust-region thresholds. In contrast, exploration tokens generate more stable sequence-level ratios, as the geometric averaging over the sequence reduces sensitivity to individual perturbations.

Formally, we treat the entire response $o=(o_1,\ldots,o_L)$ as a single action and defining its sequence-level probability as the geometric mean of the token probabilities. Consequently, we formulate $r_{\mathrm{seq\text{-}prox}}$ and $w_{\mathrm{seq\text{-}mismatch}}$:
\begin{align*}
&\pi(o\mid q) \triangleq \exp\!\left(\frac{1}{L}\sum_{t=1}^{L}\log \pi(o_t\mid q,o_{<t})\right).
\\
&r_{\mathrm{seq\text{-}prox}}(\theta) \triangleq \frac{\pi_{\mathrm{current\ learner}}(o\mid q)}{\pi_{\mathrm{old\ learner}}(o\mid q)},
\\
&w_{\mathrm{seq\text{-}mismatch}}(\theta) \triangleq \frac{\pi_{\mathrm{old\ learner}}(o\mid q)}{\pi_{\mathrm{old\ sampler}}(o\mid q)}.
\end{align*}
For mismatch weight, we truncate with a two-sided cap to control variance:
\begin{align}
\tilde{w}(\theta) \triangleq \text{clip}(w(\theta), [-\log c, \log c]),
\label{method:tis}
\end{align}
where $c>1$ is the TIS cap. For the proximal ratio, positive-advantage samples use the standard PPO-style bound $[0,\,1+\epsilon_{h}]$, while negative samples are constrained by dual-sided band $[1-\delta_{\ell},\,1+\delta_{h}]$:
\begin{align*}
\tilde r(\theta) = \begin{cases} \text{clip}(r(\theta), 0,\quad  1+\epsilon_h)& \text{if } \hat A \ge 0 \\ \text{clip}(r(\theta), 1-\delta_l, 1+\delta_h) & \text{if } \hat A < 0 \end{cases}
\label{method:clip}
\end{align*}
Put everything together, the sequence-level surrogate (shared by all tokens in $o$) is
\begin{align*}
\mathcal{J}(\theta) = \mathbb{E}_{q,\,o\sim \pi_{\mathrm{old\ sampler}}(\cdot\mid q)} \Big[ \tilde w(\theta)\cdot \tilde r(\theta)\cdot \hat A \Big].
\end{align*}

\begin{figure*}
    \centering
    \includegraphics[width=1\linewidth]{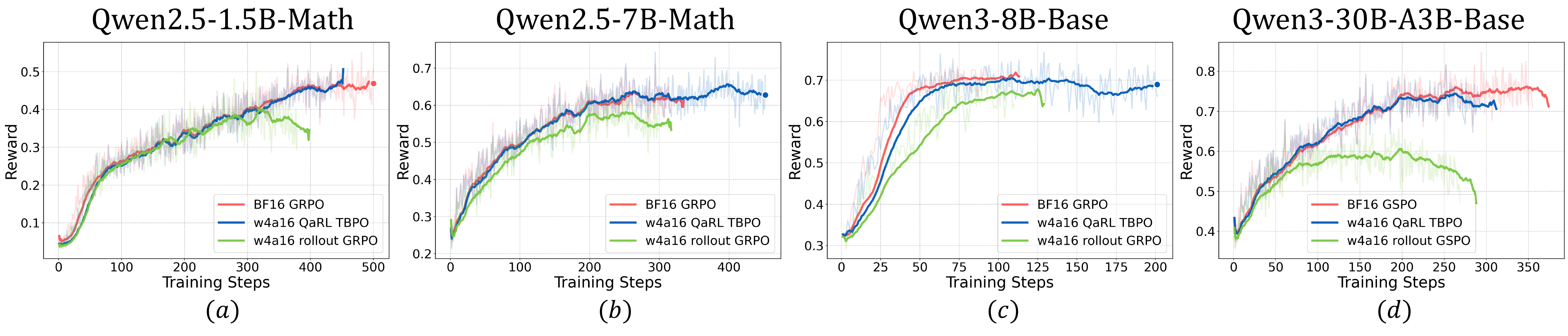}
    \vspace{-4mm}
    \caption{\textbf{Training reward curves across different models.} Our QaRL TBPO demonstrates stability over quantized rollout training, and converging to reward levels nearly identical to the full-precision BF16 baseline.}
    \label{fig:main_results}
\end{figure*}

\begin{table*}[t]
\centering
\setlength{\tabcolsep}{2.5pt}  
\renewcommand{\arraystretch}{1.3} 
\resizebox{\textwidth}{!}{
\begin{tabular}{lccccc>{\columncolor{yellow!20}}c|cccc>{\columncolor{cyan!20}}c}
\toprule
\multirow{2}{*}{\textbf{Model}} & \multicolumn{6}{c}{\textbf{In-Distribution Performance}} & \multicolumn{5}{c}{\textbf{Out-of-Distribution Performance}} \\
\cmidrule(lr){2-7} \cmidrule(lr){8-12}
 & \textbf{AIME 24/25} & \textbf{AMC} & \textbf{MATH-500} & \textbf{Minerva} & \textbf{Olympiad} & \textbf{Avg.} & \textbf{ARC-c} & \textbf{GPQA} & \textbf{MMLU-Pro} & \textbf{LiveCodeBench} &\textbf{Avg.} \\
\midrule
Qwen2.5-1.5B-Math 
  & 4.5/2.8    & 26.5 & 50.8 & 21.6  & 20.3 & 21.0      & 11.7 & 12.4 & 10.4 & 2.7 & 9.3     \\

BF16 GRPO 
  & 12.5/9.2    & 43.5 & 71.4 & 36.2  & 34.5 & 34.5      & 58.9 & 27.8 & 26.7 & 11.9 & 31.3     \\

w4a16 rollout GRPO 
  & 7.9/6.4    & 38.1 & 65.3 & 30.2  & 28.9 & 29.3      & 50.1 & 19.8 & 21.2 & 4.3 & 23.8          \\

w4a16 QaRL TBPO 
  & 12.5/10.4    & 46.6 & 69.8 & 31.9  & 32.6 & 33.9    & 57.6 & 21.1 & 25.3 & 8.1 & 28.0     \\
\midrule

Qwen2.5-7B-Math 
  & 15.0/6.4    & 46.2 & 67.4 & 32.9  & 23.9 & 33.4    & 62.6 & 28.5 & 32.1 & 8.0 & 32.8    \\

BF16 GRPO 
  & 19.5/11.6    & 59.6 & 80.0 & 45.9  & 43.4 & 43.3      & 80.4 & 38.2 & 46.3 & 14.6 & 44.8     \\

w4a16 rollout GRPO 
  & 19.1/9.6    & 54.6 & 79.1 & 42.8  & 40.6 & 40.9      & 73.9 & 31.6 & 40.0 & 8.7 & 38.5     \\

w4a16 QaRL TBPO 
  & 19.5/13.3    & 58.1 & 81.8 & 43.4  & 45.3 & 43.5    & 81.1 & 37.1 & 45.8 & 13.9 & 44.4     \\

\midrule

Qwen3-8B-Base 
  & 7.9/9.6    & 46.3 & 74.2 & 42.7  & 39.3 & 36.6    & 44.9 & 31.2 & 49.5 & 23.0 & 37.1     \\

BF16 GRPO 
  & 28.3/19.3    & 64.3 & 88.1 & 54.5  & 56.7 & 51.8      & 93.0 & 46.3 & 65.1 &45.7 & 62.5     \\

w4a16 rollout GRPO 
  & 20.0/12.5 & 53.0 & 82.2 & 50.7  & 45.1 & 43.9 &       91.6 & 43.9 & 61.5 & 41.4 & 59.5  \\

w4a16 QaRL TBPO 
  & 26.6/16.9 & 62.2 & 83.6 & 52.5  & 51.7 & 48.9       & 92.3 & 45.0 & 63.8 & 43.4 & 61.0  \\

\midrule

Qwen3-30B-A3B-Base 
  & 15.4/7.9 & 49.0 & 67.4 & 31.2  & 38.1 & 34.8       & 61.3 & 34.8 & 52.5 & 28.5 & 44.2  \\

BF16 GSPO 
  & 27.9/21.6 & 63.2 & 88.8 & 54.7  & 56.7 & 52.1       & 95.2 & 50.1 & 70.3 & 55.8 & 67.8  \\

w4a16 rollout GSPO 
  & 22.0/18.7 & 55.4 & 84.0 & 47.4  & 47.1 & 45.7       & 89.3 & 42.4 & 65.3 & 47.9 & 61.2  \\

w4a16 QaRL TBPO 
  & 27.5/22.0 & 62.9 & 87.2 & 51.4  & 56.1 & 51.2       & 96.6 & 48.2 & 68.0 & 55.4 & 67.05  \\

\bottomrule
\end{tabular}
}
\caption{Main results on in-distribution math and out-of-distribution benchmarks. LiveCodeBench results are reported in $pass@4$, while all other metrics use $pass@1$.}
\label{tab:main}
\end{table*}

\section{Experiments}
\paragraph{Training Setup.} We utilize OpenR1-Math-46K~\citep{luffy}, a dataset of 46,000 mathematical problems, and conduct experiments on Qwen2.5-Math-1.5B/7B, Qwen3-8B-Base, and Qwen3-30B-A3B-Base~\citep{yang2025qwen3}. We employ GRPO with TIS for the BF16 RL baseline and quantized training, while GSPO is adopted for MoE models to enhance stability. The Muon optimizer~\citep{jordan2024muon} exhibits significantly faster convergence than AdamW and is therefore used across all experiments; detailed hyperparameters are provided in Appendix~\ref{appendix:exp-setting}.

\paragraph{Evaluation.} We evaluate on math benchmarks (AIME2024/2025, AMC, Math-500, OlympiadBench~\citep{he2024olympiadbench}, Minerva) and out-of-distribution benchmarks (ARC-Challenge~\citep{clark2018think}, GPQA-Diamond~\citep{rein2024gpqa}, LiveCodeBench, MMLU Pro).

\begin{figure*}
    \centering    
    \includegraphics[width=1\linewidth]{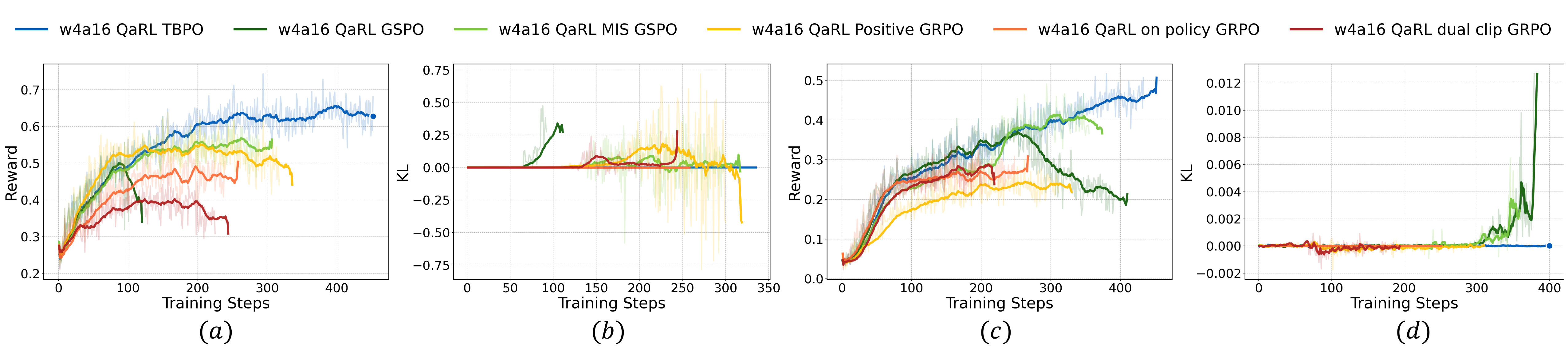}
    \caption{Training dynamics (Reward/KL) of Qwen2.5-Math 1.5B (a-b) and 7B (c-d) across different optimization objectives.}
    \label{fig:ablation}
    \vspace{-4mm}
\end{figure*}

\subsection{Main Results}
Table~\ref{tab:main} and Fig.~\ref{fig:main_results} demonstrate a consistent trend across model scales: quantized rollout training undermines stability and yields lower final accuracy than the BF16 RL baseline, whereas QaRL TBPO exhibits markedly improved stability, converging to BF16 comparable rewards. For instance, on the Qwen3-8B model, the average in-distribution math performance drops significantly from 51.8\% (BF16) to 43.9\% with quantized rollout training, while QaRL TBPO successfully maintain the performance to 48.9\%, only 2.9\% drop. Critically, QaRL TBPO recovers most of the degradation from quantized-rollout training, achieving near-baseline performance. Although quantized rollout training under GSPO still suffers from router shift at each forward pass, TBPO remains stable on MoE models, achieving an average math score of 51.2\%, nearly matching the 52.1\% baseline.

Beyond math, QaRL TBPO improves OOD performance vs quantized rollout training while matching BF16, demonstrating that gains arise from stable optimization and better generalization rather than overfitting.

\subsection{Ablation}
\begin{figure}
    \centering
    \includegraphics[width=1\linewidth]{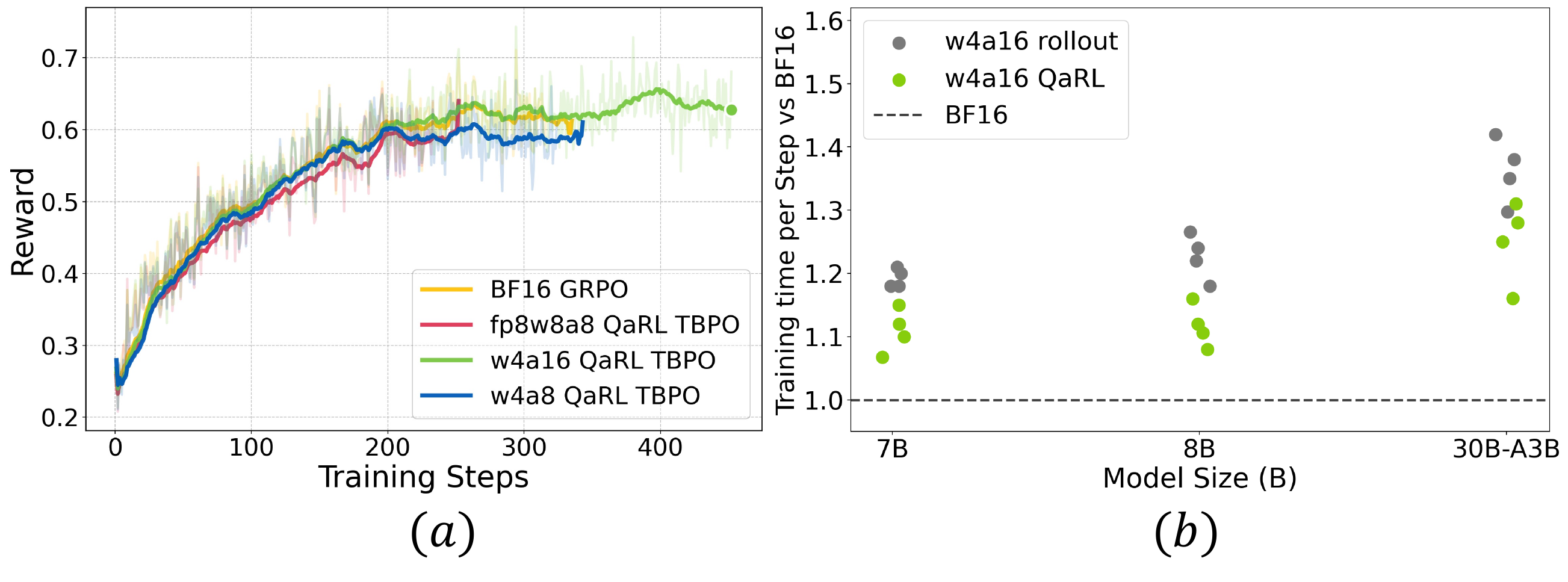}
    \caption{\textbf{(a)} Reward curves under different quant scheme. \textbf{(b)} Per-step training time speedup ratio.}
    \label{fig:quant_scheme&speed}
    \vspace{-2mm}
\end{figure}
\paragraph{Optimize objectives}
We conduct ablations to validate TBPO's effectiveness in mitigating error-token interference under QaRL (Fig.~\ref{fig:ablation}). \textbf{Overall:} TBPO achieves stable learning, high reward, and tight KL control, while alternatives suffer from KL drift or low sample efficiency. \textbf{GSPO} is unstable—error tokens in negatives cause KL drift and collapse, degrading reward. \textbf{MIS GSPO:} Rejection sampling reduces data efficiency and limits reward ceiling. \textbf{Positive GRPO:} Discarding negatives loses exploration and limits performance. \textbf{On-policy GRPO:} Single updates per batch reduce error amplification but hurt efficiency. \textbf{Dual-clip GRPO:} Clipping bounds extreme ratios, yet tokens after error tokens remain contaminated, causing incorrect learning. For comprehensive analysis of SAPO, please refer to Appendix~\ref{app:sapo}.

\paragraph{Quantization Scheme}
We further ablate quantization schemes in QaRL-TBPO, comparing FP8W8A8, W4A16, and W4A8 against the BF16 GRPO baseline (Fig.~\ref{fig:quant_scheme&speed}(a)). No substantial differences emerge: all bit-widths exhibit similar reward curves and nearly identical final rewards. This suggests that, once stabilized via TBPO, RL convergence is largely insensitive to the specific low-bit format, and performance gains are not tied to a particular quantization choice. We adopt W4A16 for most experiments due to its superior speed on larger models and broad hardware compatibility, whereas W4A8 requires more complex kernel support. Results of TBPO used on quantized rollout training are detailed in appendix~\ref{app:tbpo_ablation}.


\paragraph{Speed}
Fig.~\ref{fig:quant_scheme&speed}(b) reports per-step training time normalized to BF16. Across 7B, 8B, and 30B-A3B MoE models, both QaRL and quantized rollout training achieve speedups over BF16. For large-scale MoE models, quantized rollout training delivers a 1.4$\times$ speedup, while QaRL achieves a 1.3$\times$ speedup, since QaRL incurs modest overhead from in-training quantization. Comparison of different quantization schemes are provided in appendix~\ref{fig:app_speed}. We opted not to implement FP8 KV quantization, as KV quant currently fails to provide meaningful throughput benefits in vLLM.
\section{Related Work}

\paragraph{Quantization.}
Quantization accelerates and compresses LLMs via post-training quantization (PTQ) or quantization-aware training (QAT). PTQ methods such as GPTQ~\citep{frantar2022gptq} and AWQ~\citep{lin2024awq} avoid retraining but often struggle on reasoning models, where long CoT generation induces distribution shift and error accumulation beyond static calibration. QAT methods (e.g., LLM-QAT~\citep{liu2024llm}, EfficientQAT~\citep{chen2025efficientqat}) train with quantization noise for better robustness, while fully quantized training~\citep{wang2025infir2} executes low-bit arithmetic end-to-end to further improve robustness and throughput.

\paragraph{Reinforcement Learning in LLMs.}
Modern reasoning models (e.g., DeepSeek-R1, Qwen3~\citep{yang2025qwen3}, Kimi-K2~\citep{kimi_k2_thinking}) commonly build on GRPO. Recent variants include DAPO (higher-bound clipping to promote exploration), GSPO (sequence-level importance sampling for MoE stability), and ASPO~\citep{wang2025aspo} (asymmetric sampling for low-probability tokens). Building on this line, we focus on stabilizing RL under quantized rollouts and QAT by suppressing quantization-induced errors during generation.

\section{Conclusion and Limitations}
We propose QaRL to mitigate the severe training–inference mismatch in quantized-rollout RL via rollout-aligned training, and identify error tokens as a key driver of collapse under quantized rollouts, addressed by TBPO with a sequence-level objective and dual clipping on negative samples. Looking forward, we plan to explore fully quantized RL training, and to replace costly, low-utilization sequence-level optimization with token-level approximations that retain stability while improving efficiency and sample usage.

\bibliography{z_citation}

\appendix
\newpage
\section*{Appdenix}

\section{The Use of Large Language Models (LLMs)}
A large language model was utilized for grammatical and stylistic refinement of the manuscript. Its role was strictly limited to text editing and polishing to enhance clarity. All research ideas, experimental design, and analytical content are the original work of the authors.

\section{Experiment Setting \label{appendix:exp-setting}}
\paragraph{Training.}
We use \texttt{Verl}~\citep{sheng2024hybridflow} as the training framework and \texttt{vLLM}~\citep{kwon2023efficient} as the inference engine. All experiments are conducted on $8\times$ NVIDIA H800 GPUs.
\paragraph{Evaluation Dataset.} We evaluate our approach on standard in distribution mathematical datasets including AIME2024, AIME2025~\citep{ye2025aimepreview}, AMC, Math-500~\citep{lightman2023lets}, OlympiadBench~\citep{he2024olympiadbench} and Minerva. To further investigate the generalizability of quantized training, we extend evaluation to out-of-distribution benchmarks such a out-of-distribution benchmarks ARC-Challenge~\citep{clark2018think}, GPQA-Diamond~\citep{rein2024gpqa}, LiveCodeBench~\citep{jain2024livecodebench}, MMLU Pro~\citep{wang2024mmlu}.

\paragraph{Hyperparameters.}
\begin{itemize}[leftmargin=*, itemsep=1pt, topsep=2pt, parsep=0pt, partopsep=0pt]
    \item seq\_clip\_ratio\_high is $\epsilon_h$,
    \item neg\_seq\_clip\_ratio\_high is $\delta_h$ and neg\_seq\_clip\_ratio\_low is $\delta_l$,
    \item seq\_tis\_imp\_ratio\_cap is $c$ in equation~\ref{method:tis}.
\end{itemize}

\section{Batch invariant kernel}
\label{appendix:batch-invariant}
The root cause of output inconsistency across different kernel implementations lies in the finite precision of floating point accumulators. Due to the non associativity of floating point addition, where $(a+b)+c=a+(b+c)$ the final result is sensitive to the order of operations. To ensure bit-wise reproducibility, it is necessary to enforce a fixed summation order regardless of the batch configuration, leading to the design of a Batch-Invariant Kernel. However, imposing a strict execution order often comes at the cost of performance, as it restricts the asynchronous parallelism and dynamic scheduling inherent to GPU architectures.

\begin{table}
\centering
\begin{tabular}{l c}
\toprule
\textbf{Parameter Name}       & \textbf{Value}                    \\ 
\midrule
trainer.nnodes     &  1 \\ 
trainer.n\_gpu\_per\_node    & 8 \\
data.train\_batch\_size   & 512     \\ 
data.max\_prompt\_length    & 2048          \\ 
data.max\_response\_length   & 16384 \\
rollout.n & 8 \\
rollout.temperature & 1.0 \\
rollout.top\_p & 1.0 \\
val\_kwargs.temperature & 0.6 \\
actor.ppo\_mini\_batch\_size & 64 \\
actor.ppo\_max\_token\_len\_per\_gpu & 22528 \\
optim.opt\_type & Muon \\
optim.lr & 1e-6 \\
optim.weight\_decay & 0.01 \\
actor.use\_kl\_loss & False \\
actor.seq\_clip\_ratio\_high & 0.0004 \\
actor.seq\_clip\_ratio\_low & 0.0003 \\
actor.neg\_seq\_clip\_ratio\_high & 0.0007 \\
actor.neg\_seq\_clip\_ratio\_low & 0.0003 \\
actor.seq\_tis\_imp\_ratio\_cap & 2 \\
\bottomrule
\end{tabular}
\caption{Hyperparameters for Experiment.}
\end{table}
\section{More Experimental Results}
\subsection{Speed on different quant scheme}
Fig.~\ref{fig:app_speed} illustrates the per-step training latency of various quantization schemes on Qwen3-30B-A3B (MoE), normalized to the BF16 baseline (dashed line at 1.0). Our results show that efficiency gains become increasingly significant from W8 to W4. This trend underscores that MoE training is primarily memory/IO-bound; since MoE operators are almost inherently memory-bound during decoding, the weight bit-width directly dictates computational efficiency. Furthermore, lower precision reduces the memory footprint per model instance, minimizing the required GPU count and alleviating inter-GPU communication overhead—a benefit that becomes even more pronounced when intra-node interconnects (e.g., NVLink) are fully utilized. Consequently, we adopt W4 as the default configuration for our primary experiments.

\subsection{More Ablation on TBPO\label{app:tbpo_ablation}}
To further evaluate the effectiveness of TBPO, particularly its capability to mitigate the negative impact of "error tokens" generated by quantized rollout engines, we conduct a ablation study against GRPO under different precision settings. As demonstrated in Table~\ref{tab:tbpo_ablation}, TBPO achieves performance on par with GRPO under BF16 precision. However, under w4a16 quantized rollout training, TBPO exhibits markedly superior robustness by effectively neutralizing error token interference during optimization, thereby delivering an average margin of 2.9 points over GRPO. While TBPO provides some degree of mitigation for the mismatch issues arising from quantized rollout training, the fundamental resolution lies in leveraging QAT/QaRL or fully quantized training regimes to eliminate systemic inconsistencies at their core.

\begin{table}
    \centering
    \resizebox{0.48\textwidth}{!}{
    \begin{tabular}{lcccc}
        \toprule
        \multicolumn{5}{c}{Qwen3-8B-Base} \\
        Method & MATH-500 & AIME25 & AMC & Avg.\\ 
        \midrule
        BF16 GRPO & 88.1 & 19.3 & 64.3 & 57.2\\
        BF16 TBPO & 86.3 & 20.0 & 62.4 & 56.2\\
        w4a16 GRPO & 82.2 & 12.5 & 53.0 & 49.2\\
        w4a16 TBPO & 84.9 & 13.3 & 58.3 & 52.1\\
        \bottomrule
    \end{tabular}
    }
    \caption{Performance comparison between GRPO and TBPO under full-precision (BF16) and quantized (w4a16) rollout training settings.}
    \label{tab:tbpo_ablation}
\end{table}

\begin{figure}
    \centering
    \includegraphics[width=0.9\linewidth]{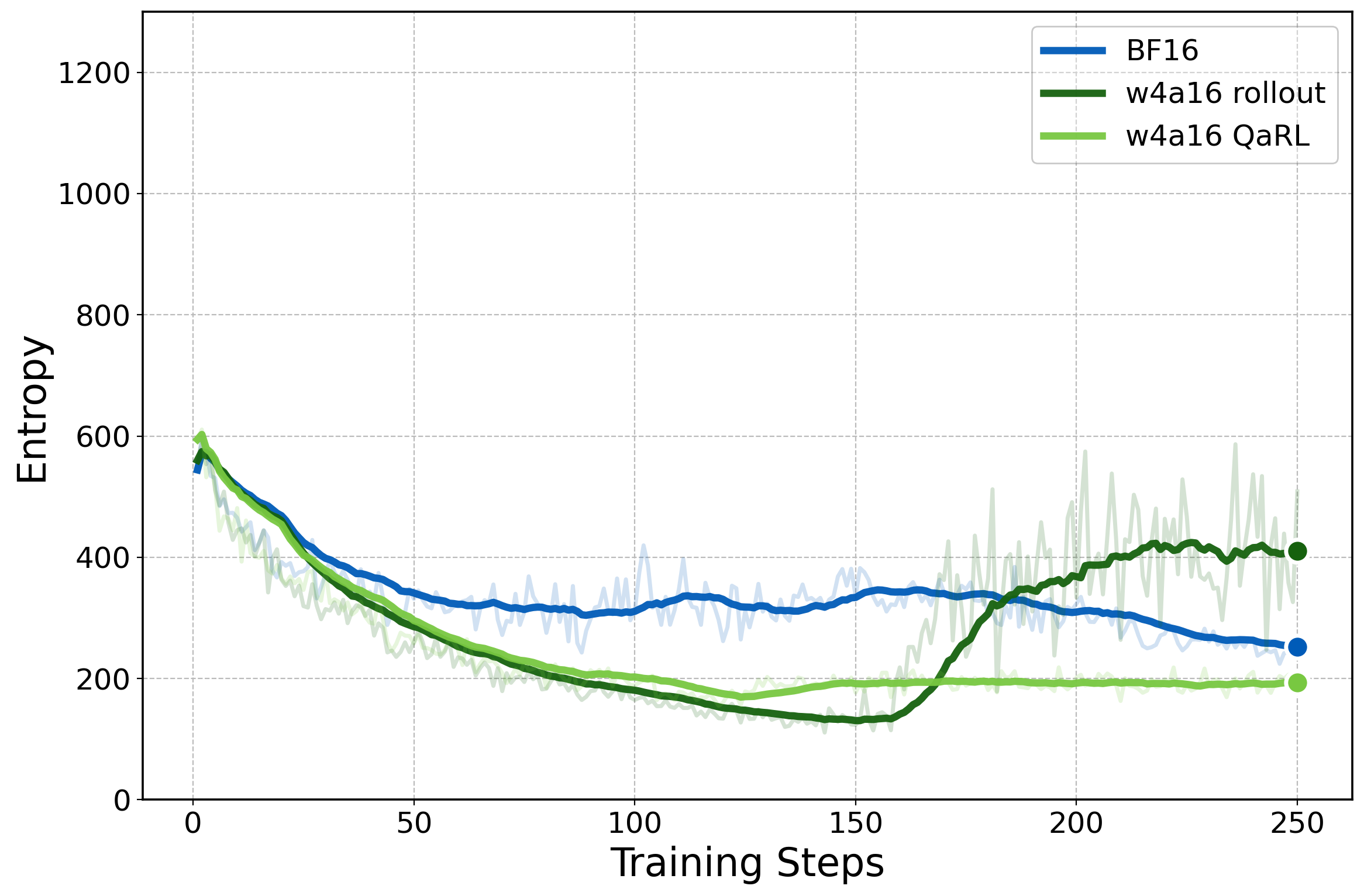}
    \caption{Comparison of RL training entropy}
    \label{fig:app_entropy}
\end{figure}
\begin{figure}
    \centering
    \includegraphics[width=0.9\linewidth]{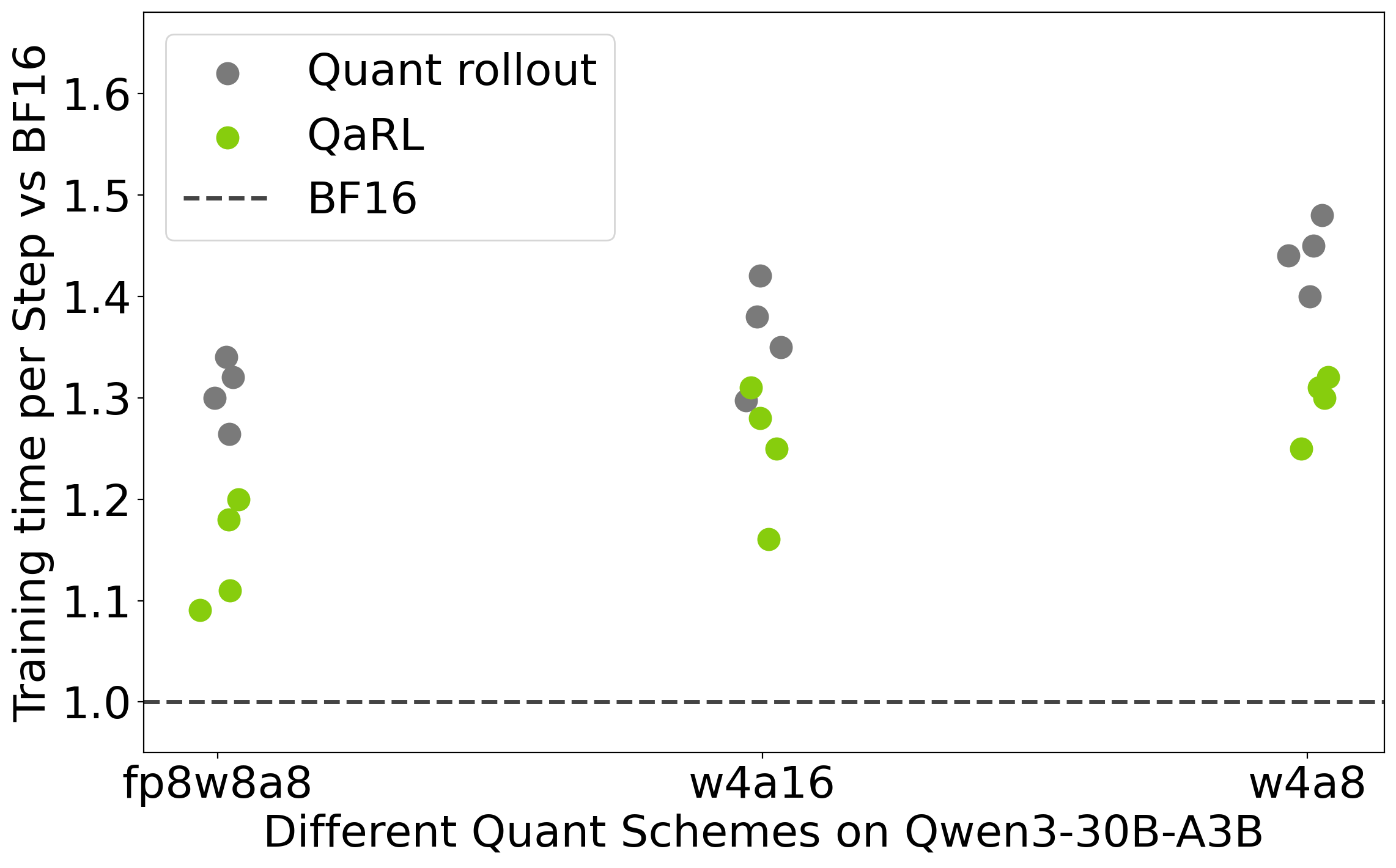}
    \caption{Different quant scheme speed on Qwen3-30B-A3B MoE model.}
    \label{fig:app_speed}
\end{figure}
\subsection{Entropy of quantized RL training\label{app:entropy}}
Whereas~\citet{huang2025qerl} Fig.~5 posits that quantization errors during training may attenuate the entropy reduction in RL, our empirical findings present a more nuanced picture. Across both quantized rollout training and QAT/QaRL paradigms (Fig.~\ref{fig:app_entropy}), quantization discrepancies are progressively assimilated throughout the optimization process, yielding no discernible entropy elevation relative to the BF16 baseline. The entropy escalation observed in the latter stages of w4a16 quantized rollout training stems from the over-optimization of error tokens, which precipitates repetitive generation patterns.

\begin{figure}
    \centering
    \includegraphics[width=0.9\linewidth]{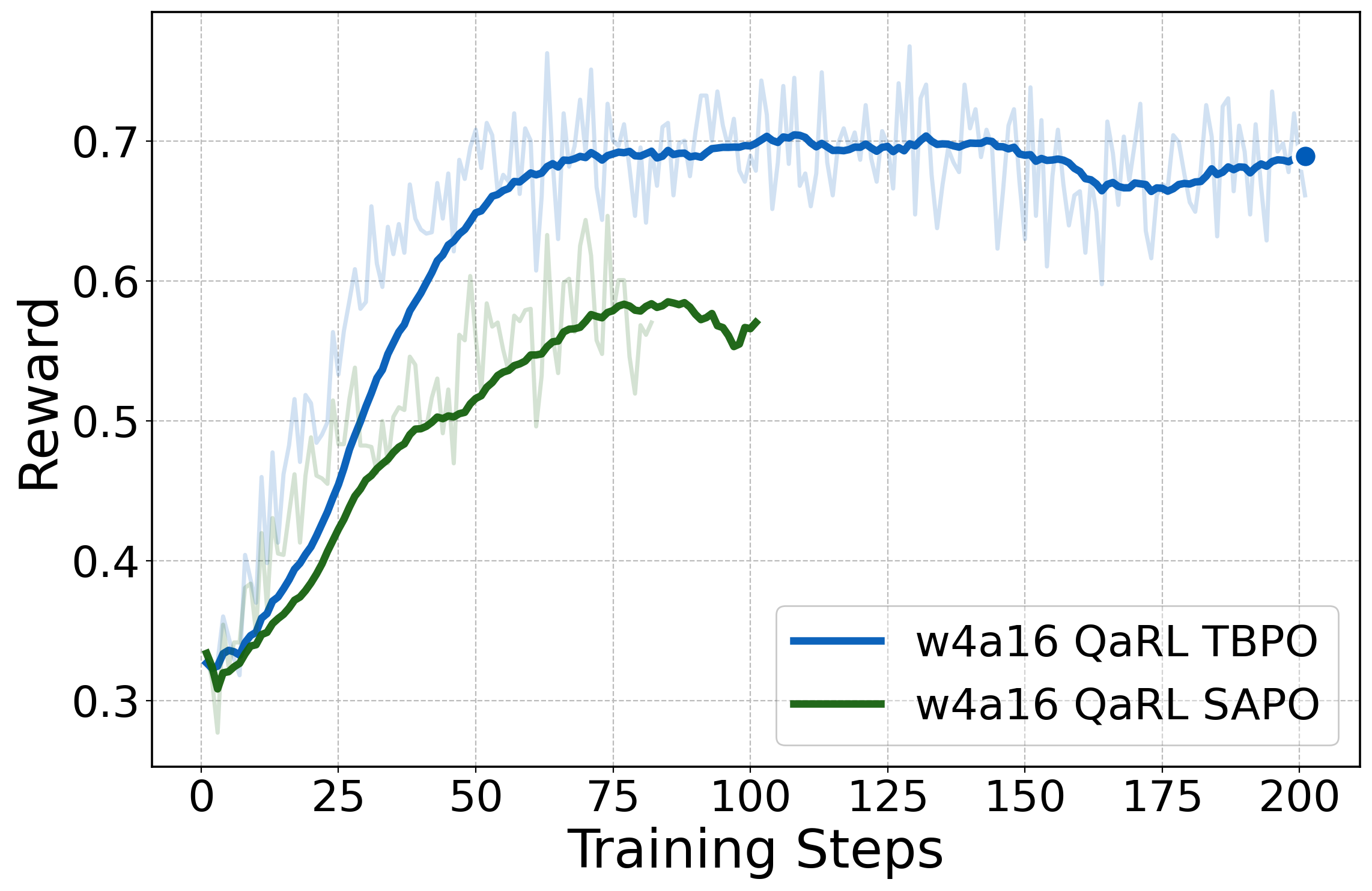}
    \caption{Comparison of SAPO on QaRL}
    \label{fig:sapo}
\end{figure}
\subsection{Comparsion of soft weighted methods SAPO\label{app:sapo}}
SAPO~\citep{gao2025soft} introduces a soft adaptive weighting mechanism as a nuanced alternative to hard clipping, aiming to bolster optimization stability in MoE architectures. Through dynamic reweighting of token contributions—assigning diminishing importance to tokens whose advantage ratios deviate further from unity—SAPO adeptly discriminates between erroneous tokens (demanding substantial correction) and exploration tokens (that foster policy discovery). Nevertheless, as evidenced in Fig.~\ref{fig:sapo}, SAPO remains unable to surpass TBPO in the context of quantized RL training. We posit that this suboptimal performance stems from SAPO's continued assimilation of responses containing error tokens, despite attenuating their individual contributions to policy updates. Critically, such response sequences are globally off-distribution, rendering them intrinsically unsuitable for stable policy learning. This observation reaffirms the paramount importance of constraining sequence-level optimization objectives within the trust region, while simultaneously attesting to the elegance and principled simplicity inherent in TBPO's design.

\begin{figure*}
\begin{align*}
&r_{\mathrm{seq\text{-}prox}}(\theta) \triangleq \frac{\pi_{\mathrm{current\ learner}}(o\mid q)}{\pi_{\mathrm{old\ learner}}(o\mid q)} = \exp\!\left(\frac{1}{L}\sum_{t=1}^{L} (\log\pi_{\mathrm{current\ learner}}-\log\pi_{\mathrm{old\ learner}})(o_t\mid q,o_{<t})\right),
\\
&w_{\mathrm{seq\text{-}mismatch}}(\theta) \triangleq \frac{\pi_{\mathrm{old\ learner}}(o\mid q)}{\pi_{\mathrm{old\ sampler}}(o\mid q)} = \exp\!\left(\frac{1}{L}\sum_{t=1}^{L} (\log\pi_{\mathrm{old\ learner}}-\log\pi_{\mathrm{old\ sampler}})(o_t\mid q,o_{<t})\right).
\end{align*}
\caption{Sequence ratio and weight definition.\label{fig:define}}
\end{figure*}
\section{Detail of TBPO}
To effectively optimize the policy while maintaining training stability, we define the sequence-level importance weights and mismatch factors. Specifically, for a given query $q$ and its corresponding output sequence $o$ of length $L$, we introduce the sequence-level proximity ratio $r_\text{seq-prox}(\theta)$ and the sequence-level mismatch weight $w_\text{seq-mismatch}(\theta)$ as illustrated in Fig.~\ref{fig:define}.

Instead of using a simple product of token-level probabilities, which often leads to vanishing or exploding gradients in long-context scenarios, we employ the geometric mean of token-level ratios (formulated as the exponential of the average log-difference). This design ensures that the importance weights remain numerically stable across varying sequence lengths. $r_\text{seq-prox}(\theta)$ serves to constrain the policy update within a reliable trust region by measuring the drift from the previous learner, while $w_\text{seq-mismatch}(\theta)$ accounts for the distributional shift between the historical sampling policy and the current optimization baseline. 

\begin{table*}

\begin{box3}{GRPO token level clipping \& weighting}
Response1: \hltoken{green4}{Use quantized}
\hltoken{green3}{rollout }
\hltoken{green1}{engine}
\hltoken{clippedGray}{\sout{to to to to to to to to to to to to to }}
\hltoken{green2}{accelerate RL}
\notedtoken{green3}{wide sees gr }{1}
\hltoken{clippedGray}{\sout{ContributionsĠÐ±ÐµÐ}}
\\
Response2: \hltoken{green1}{Train}
\hltoken{green3}{-inference mismatch}
\hltoken{green2}{is a}
\hltoken{green4}{primary cause}
\hltoken{green1}{of training collapse in RL}
\\
Response3: \notedtoken{clippedGray}{\sout{Aha}{ }}{2}
\hltoken{green1}{! We can }
\hltoken{green4}{clip error token to}
\hltoken{clippedGray}{\sout{keep}}
\hltoken{green1}{optimization in trust region}
\end{box3}

\begin{box3}{TBPO sequence level clipping \& weighting}
Response1: \hltoken{clippedGray}{\sout{Use quantized rollout engineto to to to to to to to to to to to to to accelerate RL}}
\hltoken{clippedGray}{\sout{wide sees gr ContributionsĠÐ±ÐµÐ}}

Response2: \hltoken{green1}{Train-inference mismatch is a primary cause of training collapse in RL}

Response3: \hltoken{green2}{Aha! We can clip error token to keep optimization in trust region}
\end{box3}
\caption{Comparison of clipping and weighting strategies. \hltoken{clippedGray}{Gray} highlights clipped tokens, while \hltoken{green2}{different colors} represent varying weighting magnitudes (darker shades indicate weights further from 1). GRPO employs token-level clipping and weighting, whereas TBPO utilizes sequence-level granularity. In GRPO, the garbled tokens (marked by 1) are not clipped, while the exploration tokens (marked by 2) are clipped. In contrast, TBPO's sequence-level approach correctly clips the garbled tokens without erroneously clipping the exploration tokens.}
\label{app:tokens}
\end{table*}

\end{document}